\documentclass{article}

\usepackage{microtype}
\usepackage{graphicx}
\usepackage{subfigure}
\usepackage{booktabs} 
\usepackage{soul}

\usepackage{hyperref}

\newcommand\biasamptheirs{{{$\textnormal{BiasAmp}_{\mathrm{MALS}}$}}}
\newcommand\biasampours{{{$\textnormal{BiasAmp}_{\rightarrow}$}}}
\newcommand\biasampoursTA{{{$\textnormal{BiasAmp}_{T\rightarrow A}$}}}
\newcommand\biasampoursAT{{{$\textnormal{BiasAmp}_{A \rightarrow T}$}}}

\newcommand{\mathbbm}[1]{\text{\usefont{U}{bbm}{m}{n}#1}} 
\newcommand{\smallsec}[1]{\vspace{0cm} \noindent {\bf #1.}}

\usepackage{amsmath}
\usepackage{array}
\usepackage{enumitem}
\usepackage{multirow}
\usepackage{longtable}

\usepackage[accepted]{icml2021}

\icmltitlerunning{Directional Bias Amplification}

\begin{document}

\twocolumn[
\icmltitle{Directional Bias Amplification}



\icmlsetsymbol{equal}{*}

\begin{icmlauthorlist}
\icmlauthor{Angelina Wang and Olga Russakovsky}{}
\end{icmlauthorlist}

\centering Princeton University


\icmlcorrespondingauthor{Angelina Wang}{angelina.wang@princeton.edu}

\icmlkeywords{bias amplification, fairness, evaluation}

\vskip 0.3in
]



\printAffiliationsAndNotice{}  

\begin{abstract}

Mitigating bias in machine learning systems requires refining our understanding of bias propagation pathways: 
from societal structures to large-scale data to trained models to impact on society. In this work, we focus on one aspect of the problem, namely \emph{bias amplification}: the tendency of models to amplify the biases present in the data they are trained on. A metric for measuring bias amplification was introduced in the seminal work by \citet{zhao2017menshop}; however, as we demonstrate, this metric suffers from a number of shortcomings including conflating different types of bias amplification and failing to account for varying base rates of protected attributes. We introduce and analyze a new, decoupled metric for measuring bias amplification, \biasampours{} (Directional Bias Amplification). We thoroughly analyze and discuss both the technical assumptions and normative implications of this metric. We provide suggestions about its measurement by cautioning against predicting sensitive attributes, encouraging the use of confidence intervals due to fluctuations in the fairness of models across runs, and discussing the limitations of what this metric captures.
Throughout this paper, we work to provide an interrogative look at the technical measurement of bias amplification, guided by our normative ideas of what we want it to encompass. Code is located at \url{https://github.com/princetonvisualai/directional-bias-amp}.

\end{abstract}

\section{Introduction}
The machine learning community is becoming increasingly cognizant of problems surrounding fairness and bias, and correspondingly, a plethora of new algorithms and metrics are being proposed (see e.g., \citet{mehrabi2019survey} for a survey). 
The analytic gatekeepers of the systems often take the form of fairness evaluation metrics, and it is vital that these be deeply investigated both technically and normatively. In this paper, we endeavor to do this for bias amplification.

Bias amplification occurs when a model exacerbates biases from the training data at test time. It is the result of the algorithm~\citep{foulds2018intersectionality}, and unlike some other forms of bias, cannot be solely attributed to the dataset. 




\begin{figure*}[t]
\includegraphics[width=0.95\linewidth]{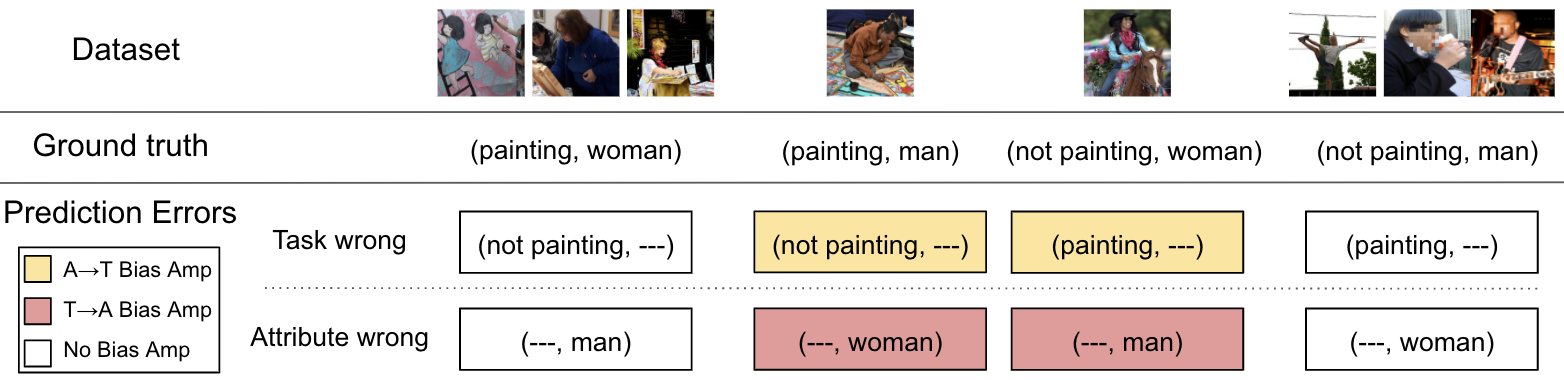} 
\caption{\label{fig:fig1}Consider an image dataset where the goal is to classify the task, $T$, as painting or not painting, and the attribute, $A$, as woman or man. Women are correlated with painting, and men with not painting. In this work we are particularly concerned with errors contributing to the amplification of bias, i.e., existing training correlations (yellow and red in the figure). We further disentangle these errors into those that amplify the attribute to task correlation (i.e., incorrectly predict the task based on the person's attribute; shown in yellow) versus those that amplify the task to attribute (shown in red).}
\end{figure*}

\smallsec{Directional bias amplification metric}
We propose a new way of measuring bias amplification, \biasampours{} (Directional Bias Amplification),\footnote{The arrow is meant to signify the direction that bias amplification is flowing, and not intended to be a claim about causality.} that builds off a prior metric from ``Men Also Like Shopping: Reducing Gender Bias Amplification using Corpus-level Constraints''~\citep{zhao2017menshop}, that we will call \biasamptheirs{}. Our metric's technical composition aligns with the real-world qualities we want it to encompass, addressing a number of the previous metric's shortcomings by being able to: 1) focus on both positive and negative correlations, 2) take into account the base rates of each protected attribute, and most importantly 3) disentangle the directions of amplification.

As an example, consider a visual dataset (Fig.~\ref{fig:fig1}) where each image has a label for the task $T$, which is painting or not painting, and further is associated with a protected attribute $A$, which is woman or man.\footnote{We use the terms \emph{man} and \emph{woman} to refer to binarized socially-perceived (frequently annotator-inferred) gender expression, recognizing these labels are not inclusive and may be inaccurate.} If the gender of the person biases the prediction of the task, we consider this $A\rightarrow T$ bias amplification; if the reverse happens, then $T\rightarrow A$. Bias amplification as it is currently being measured merges together these two different paths which have different normative implications and therefore demand different remedies. 
This speaks to a larger problem of imprecision when discussing problems of bias~\cite{blodgett2020nlpbias}. For example, ``gender bias" can be vague; it is unclear if the system is assigning gender in a biased way, or if there is a disparity in model performance between different genders. 
Both are harmful in different ways, but the conflation of these biases can lead to misdirected solutions.

\smallsec{Bias amplification analysis} The notion of bias amplification allows us to encapsulate the idea that systemic harms and biases can be more harmful than errors made without such a history~\citep{bearman09sexism}. For example, in images, overclassifying women as cooking carries a more negative connotation than overclassifying men as cooking. The distinction of which errors are more harmful can often be determined by lifting the patterns from the training data.

In our analysis and normative discussion, we look into this and other implications through a series of experiments. We consider whether predicting protected attributes is necessary in the first place; by not doing so, we can trivially remove $T\rightarrow A$ amplification. We also encourage the use of confidence intervals because
\biasampours{}, along with other fairness metrics, suffers from the Rashomon Effect~\citep{breiman2001cultures}, or multiplicity of good models. 
In other words, in supervised machine learning, random seeds have relatively little impact on \emph{accuracy}; in contrast, they appear to have a greater impact on fairness.

Notably, a trait of bias amplification is that it is not at odds with accuracy. Bias amplification measures the model's errors, so a model with perfect accuracy will have perfect (zero) bias amplification. (Note nevertheless that the metrics are not always correlated.) 
This differs from many other fairness metrics, because the goal of not amplifying biases and thus matching task-attribute correlations is aligned with that of accurate predictions.
For example, satisfying fairness metrics like demographic parity~\citep{dwork2012awareness} are incompatible with perfect accuracy when parity is not met in the ground-truth.
For the same reason bias amplification permits a classifier with perfect accuracy, it also comes with a set of limitations that are associated with treating data correlations as the desired ground-truth, and thus make it less appropriate for social applications where other metrics are better suited for measuring a fair allocation of resources.



\smallsec{Outline} To ground our work, we first distinguish what bias amplification captures that standard fairness metrics cannot, then distinguish \biasampours{} from \biasamptheirs{}. Our key contributions are: 1) proposing a new way to measure bias amplification, addressing multiple shortcomings of prior work and allowing us to better diagnose models, and 2) providing a comprehensive technical analysis and normative discussion around the use of this measure in diverse settings, encouraging thoughtfulness with each application. 




\section{Related Work}
\smallsec{Fairness Measurements}
Fairness is nebulous and context-dependent, and approaches to quantifying it include equalized odds~\citep{hardt2016equalopp}, equal opportunity~\citep{hardt2016equalopp}, demographic parity~\citep{dwork2012awareness, kusner2017counterfactual}, fairness through awareness~\citep{dwork2012awareness, kusner2017counterfactual}, fairness through unawareness~\citep{hlaca2016case, kusner2017counterfactual}, and treatment equality~\citep{berk2017risk}.
We examine bias amplification, a type of group fairness where correlations are amplified.



\begin{figure}[t]
\centering
\begin{subfigure}{}
\includegraphics[width=.9\linewidth]{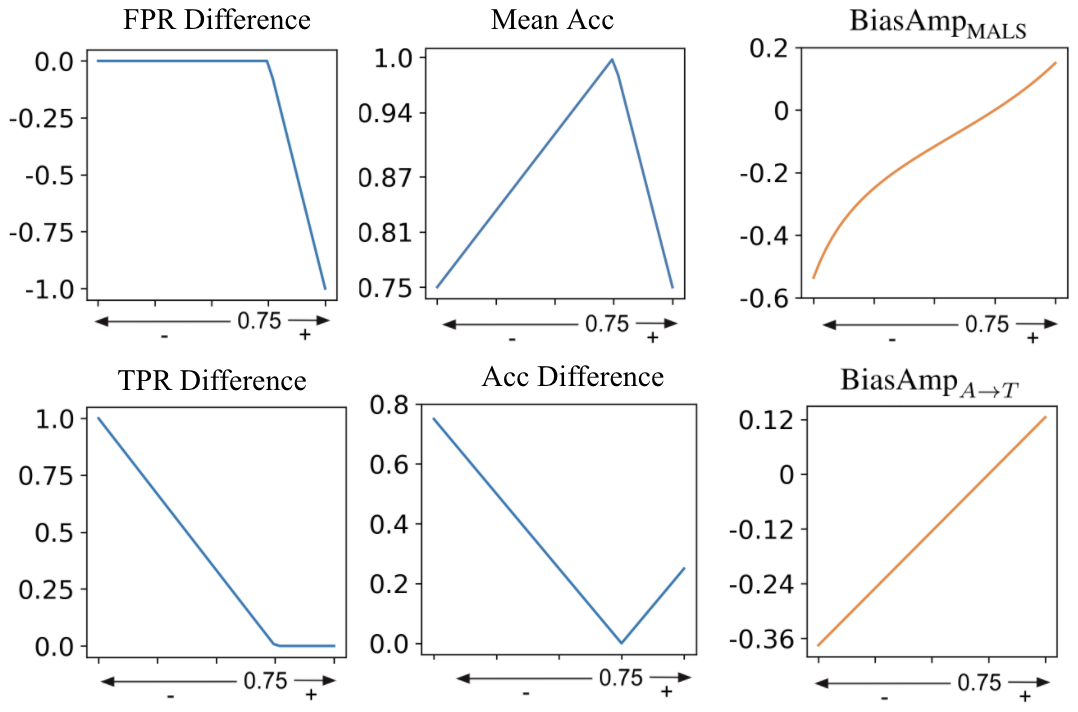} 
\end{subfigure}
\caption{\label{fig:metric_comp}
Fairness metrics vary in how they respond to model errors. In our image dataset (Fig.~\ref{fig:fig1}) of predicting someone who is a woman or man to be painting or not, we consider a painting classifier that always predicts the task correctly for men, but varies for women. The x-axes correspond to the percentage of women predicted to be painting, where the ground-truth is 0.75. Below that the model is under-predicting women to be painting, and above it the model is over-predicting.
The two metrics in the first column, FPR and TPR difference, only capture one of under- or over-prediction. The next two metrics in the second column, accuracy difference between attribute subgroups and average mean-per-class accuracy across attribute subgroups, are symmetric around 0.75, being unable to differentiate. Thus, the bias amplification metrics in the last column are needed to distinguish between under- and over-prediction (\biasamptheirs{} from \citet{zhao2017menshop} in Sec.~\ref{sec:biasamptheirs-shortcomings}, and our proposed \biasampoursAT{} in Sec.~\ref{sec:our_metrix_exp}). \biasamptheirs{} requires attribute predictions, so we assume perfect attribute prediction here to make the comparison.}
\end{figure}

As an example of what differentiates bias amplification, we present a scenario based on Fig.~\ref{fig:fig1}. We want to classify a person whose attribute is man or woman with the task of painting or not. The majority groups ``(painting, woman)" and ``(not painting, man)" each have 30 examples, and the minority groups ``(not painting, woman)" and ``(painting, man)" each have 10. A classifier trained to recognize painting on this data is likely to learn these associations and over-predict painting on images of women and under-predict painting on images of men; however, algorithmic interventions may counteract this and result in the opposite behavior. In Fig.~\ref{fig:metric_comp} we show how four standard fairness metrics (in blue) vary under different amounts of learned amplification: FPR difference, TPR difference~\citep{chouldechova2016recidivism, hardt2016equalopp}, accuracy difference in task prediction~\citep{berk2017risk}, and average mean-per-class accuracy across subgroups~\citep{buolamwini2018gendershades}. 
However, these four metrics are not designed to account for the training correlations, and unable to distinguish between cases of increased or decreased learned correlations, motivating a need for a measurement that can: bias amplification.

\smallsec{Bias Amplification}
Bias amplification has been measured by looking at binary classifications without attributes~\citep{leino2019feature}, GANs~\citep{jain2020gan, choi2020gan}, and correlations~\citep{zhao2017menshop, jia2020posterior}. We consider attributes in our formulation, which is a classification setting, and thus differs from GANs. We dissect in detail the correlation work, and propose measuring \emph{conditional} correlations, which we term ``directional.” \citet{wang2019balanced} measures amplification by predicting the sensitive attribute from the model outputs, thus relying on multiple target labels simultaneously; we propose a decomposable metric to allow for more precise model diagnosis.

The Word Embedding Association Test (WEAT)~\citep{caliskan2017weat} measures bias amplification in de-contextualized word embeddings, specifically, non-conditional correlations
~\citep{bolukbasi2016embeddings}. However, with newer models like BERT and ELMo that have contextualized embeddings, WEAT does not work~\citep{may2019sentence}, so new techniques have been proposed incorporating context~\citep{lu2019nlp, kuang2016semantic}. We use these models to measure the directional aspect of amplifications, as well as to situate them in the broader world of bias amplification. 
Directionality of amplification has been observed~\citep{stock2018mistakes, qian2019genderequalizing}, but
we take a more systematic approach.


\smallsec{Causality} Bias amplification is also studied in the causal statistics literature~\citep{bhattacharya2007, wooldridge2016, pearl2010, pearl2011, middleton2016}. However, despite the same terminology, the definitions and implications are largely distinct. Our work follows the machine learning bias amplification literature discussed in the previous section and focuses on the amplification of socially-relevant correlations in the training data.

\smallsec{Predictive Multiplicity}
The Rashomon Effect~\citep{breiman2001cultures}, or multiplicity of good models, has been studied in various contexts. The variables investigated that differ across good models include explanations~\citep{li2020explanations}, individual treatments~\citep{marx2020multiplicity, pawelczyk2020multiplicity}, and variable importance~\citep{fisher2019wrong, dong2019clouds}. We build on these and investigate how fairness also differs between equally ``good" models. 

\section{Existing Bias Amplification Metric}


We describe the existing metric~\cite{zhao2017menshop} and highlight shortcomings that we address in Sec.~\ref{sec:our_metrix_exp}. 

\subsection{Notation}
\label{sec:biasamptheirs-notation}

Let $\mathcal{A}$ be the set of protected demographic groups: for example, $\mathcal{A} = \{$woman, man$\}$ in Fig.~\ref{fig:fig1}. $A_a$ for $a\in \mathcal{A}$ is the binary random variable corresponding to the presence of the group $a$; thus $P(A_{\texttt{woman}}= 1)$ can be empirically estimated as the fraction of images in the dataset containing women. Note that this formulation is generic enough to allow for multiple protected attributes and intersecting protected groups. Let $T_t$ with $t\in\mathcal{T}$ similarly correspond to binary target tasks, e.g., $\mathcal{T}= \{$painting$\}$ in Fig.~\ref{fig:fig1}. 



\subsection{Formulation and shortcomings}
\label{sec:biasamptheirs-shortcomings}

Using this notation, \citet{zhao2017menshop}'s metric is:
\begin{align}
\label{eqn:menshop}
\text{\biasamptheirs}&=
        \frac{1}{|\mathcal{T}|}\sum_{a\in \mathcal{A}, t\in\mathcal{T}}y_{at}\Delta_{at}\\
        \text{with }  y_{at} &= \mathbbm{1}\left[ P(A_a=1|T_t=1) > \frac{1}{|\mathcal{A}|} \right ]
                    \notag \\
          \Delta_{at} &= P(\hat{A}_a=1|\hat{T}_t=1) - P(A_a=1|T_t=1) \notag
\end{align}
where $\hat{A}_a$ and $\hat{T}_t$ denote model predictions for the protected group $a$ and the target task $t$ respectively.
One attractive property of this metric is that it doesn't require any ground truth test labels: assuming the training and test distributions are the same, $P(A_a=1|T_t=1)$ can be estimated on the training set, and $P(\hat{A}_a=1|\hat{T}_t=1)$ relies only on the \emph{predicted} test labels. However, it also has a number of  shortcomings.\

\smallsec{Shortcoming 1: The metric focuses only on \emph{positive} correlations} This may lead to numerical inconsistencies, especially in cases with multiple protected groups. 

To illustrate, consider a scenario with 3 protected groups $A_1$, $A_2$, and $A_3$ (disjoint; every person belongs to one), one binary task $T$, and the following dataset\footnote{For the rest of this subsection, for simplicity since we have only one task, we drop the subscript $t$ so that $T_t$, $y_{at}$ and $\Delta_{at}$ become $T$, $y_a$ and $\Delta_a$ respectively. Further, assume the training and test datasets have the same number of examples (exs.).} 
:
\begin{itemize}[noitemsep,topsep=0pt]
    \item When $A_1=1$: {\bf 10} exs. of $T=0$ and {\bf 40} exs. of $T=1$
    \item When $A_2=1$: {\bf 40} exs. of $T=0$ and {\bf 10} exs. of $T=1$
    \item When $A_3=1$: {\bf 10} exs. of $T=0$ and {\bf 20} exs. of $T=1$
\end{itemize}

From Eq.~\ref{eqn:menshop}, we see $y_1=1$, $y_2=0$, $y_3=0$. Now consider a model that always makes correct predictions of the protected attribute $\hat{A}_a$, always correctly predicts the target task when $A_1=1$, but predicts $\hat{T}=0$ whenever $A_2=1$ and $\hat{T}=1$ whenever $A_3=1$. Intuitively, this would correspond to a case of overall learned bias amplification. However, Eqn.~\ref{eqn:menshop} would measure bias amplification as 0 since the strongest positive correlation (in the $A_1=1$ group) is not amplified.

Note that this issue doesn't arise as prominently when there are only 2 disjoint protected groups (binary attributes), which was the case implicitly considered in \citet{zhao2017menshop}. However, even with two groups there are miscalibration concerns. For example, consider the dataset above but only with the $A_1=1$ and $A_2=1$ examples. A model that correctly predicts the protected attribute $\hat{A}_a$, correctly predicts the task on $A_1=1$, yet predicts $\hat{T}=0$ whenever $A_2=1$ would have bias amplification value of $\Delta_1 = \frac{40}{40}-\frac{40}{50} = 0.2$. However, a similar model that now correctly predicts the task on  $A_2=1$ but always predicts $\hat{T}=1$ on $A_1=1$ would have a much smaller bias amplification value of $\Delta_1 = \frac{50}{60}-\frac{40}{50} = 0.033$, although intuitively the amount of bias amplification is the same. 

\smallsec{Shortcoming 2: The chosen protected group may not be \emph{correct} due to imbalance between groups} To illustrate, consider a scenario with 2 disjoint protected groups:
\begin{itemize}[noitemsep,topsep=0pt]
    \item When $A_1=1$: {\bf 60} exs. of $T=0$ and {\bf 30} exs. of $T=1$
    \item When $A_2=1$: {\bf 10} exs. of $T=0$ and {\bf 20} exs. of $T=1$
\end{itemize}
We calculate $y_1 = \mathbbm{1}\left[\frac{30}{50} > \frac{1}{2} \right ] = 1$ and $y_2 = 0$, even though the correlation is actually the reverse. Now a model, which always predicts $\hat{A}_a$ correctly, but intuitively amplifies bias by predicting $\hat{T}=0$ whenever $A_1=1$ and predicting $\hat{T}=1$ whenever $A_2=1$ would actually get a \emph{negative} bias amplification score of $\frac{0}{30}-\frac{30}{50} = -0.6$. 

\biasamptheirs{} erroneously focuses on the protected group with the most examples ($A_1=1$) rather than on the protected group that is actually correlated with $T=1$ ($A_2=1$). This situation manifests when $\min\left(\frac{1}{|\mathcal{A}|}, P(A_a=1)\right)<P(A_a=1|T_t=1)<\max\left(\frac{1}{|\mathcal{A}|}, P(A_a=1)\right)$, which is more likely to arise as the the distribution of attribute $A_a=1$ becomes more skewed.

\smallsec{Shortcoming 3: The metric entangles \emph{directions} of bias amplification} By considering only the predictions rather than the ground truth labels at test time, we are unable to distinguish between errors stemming from $\hat{A}_a$ and those from $\hat{T}$. For example, looking at just the test predictions we may know that the prediction pair $\hat{T}=1,\hat{A_1}=1$ is overrepresented, but do not know whether this is due to overpredicting  $\hat{T}=1$ on images with $A_1=1$ or vice versa. 

\subsection{Experimental analysis}
\label{sec:biasamptheirs-COCO}

To verify that the above shortcomings manifest in practical settings, we revisit the analysis of \citet{zhao2017menshop} on the COCO~\cite{lin14coco} image dataset with two disjoint protected groups $A_{\texttt{woman}}$ and $A_{\texttt{man}}$, and 66 binary target tasks, $T_t$, corresponding to the presence of 66 objects in the images. We directly use the released model predictions of $\hat{A}_a$ and $\hat{T}_t$ from \citet{zhao2017menshop}.

First, we observe that in COCO there are about 2.5x as many men as women, leading to shortcoming 2 above. Consider the object \texttt{oven}; \biasamptheirs{} calculates
$ P(A_{\texttt{man}}=1|T_{\texttt{oven}}=1)=0.56>\frac{1}{2}$
and thus considers this to be correlated with men rather than women. However, computing $P(A_{\texttt{man}}=1,T_{\texttt{oven}}=1) =  0.0103 < 0.0129 = P(A_{\texttt{man}}=1)P(T_{\texttt{oven}}=1)$
reveals that men are in fact \emph{not} correlated with oven, and this result stems from the fact that men are overrepresented in the dataset generally. Not surprisingly, the model trained on this data associates women with ovens and underpredicts men with ovens at test time, i.e., $P(\hat{A}_{\texttt{man}}=1 | \hat{T}_{\texttt{oven}}=1) -  P(A_{\texttt{man}}=1 | T_{\texttt{oven}}=1) = -0.10$, erroneously measuring \emph{negative} bias amplification. 

In terms of directions of bias amplification, we recall that  \citet{zhao2017menshop} discovers that ``Technology oriented categories
initially biased toward men such as \texttt{keyboard}... have each increased their bias toward
males by over 0.100." Concretely, from Eqn.~\ref{eqn:menshop}, $P(A_{\texttt{man}}=1|T_{\texttt{keyboard}}=1) = .70$ and $P(\hat{A}_{\texttt{man}}=1|\hat{T}_{\texttt{keyboard}}=1) = .83$, demonstrating an amplification of bias. However, the \emph{direction} or cause of bias amplification remains unclear: is the presence of man in the image increasing the probability of predicting a keyboard, or vice versa? Looking more closely at the model's disentangled predictions, we see that when conditioning on the attribute, $P(\hat{T}_{\texttt{keyboard}}=1|A_{\texttt{man}}=1)  = 0.0020 < 0.0032 =  P(T_{\texttt{keyboard}}=1|A_{\texttt{man}}=1)$, and when conditioning on the task, 
$P(\hat{A}_{\texttt{man}}=1|T_{\texttt{keyboard}}=1)  = 0.78 > 0.70 =  P(A_{\texttt{man}}=1|T_{\texttt{keyboard}}=1)$,
indicating that while keyboards are under-predicted on images with men, men are over-predicted on images with keyboards. Thus the root cause of this amplification appears to be in the gender predictor rather than the object detector. Such disentangement allows us to properly focus algorithmic intervention efforts.

Finally, we make one last observation regarding the results of \citet{zhao2017menshop}. The overall bias amplification is measured to be $.040$. However, we observe that ``man" is being predicted at a higher rate (75.6\%) than is actually present (71.2\%). With this insight, we tune the decision threshold on the validation set such that the gender predictor is well-calibrated to be predicting the same percentage of images to have men as the dataset actually has. When we calculate \biasamptheirs{} on these newly thresholded predictions for the test set, we see bias amplification drop from $0.040$ to $0.001$ just as a result of this threshold change, outperforming even the solution proposed in \citet{zhao2017menshop} of corpus-level constraints, which achieved a drop to only $0.021$. Fairness can be quite sensitive to the threshold chosen~\citep{chen2020threshold}, so careful threshold selection should be done, rather than using a default of $0.5$.
When a threshold is needed in our experiments, we pick it to be well-calibrated on the validation set. In other words, we estimate the expected proportion $p$ of positive labels from the training set and choose a threshold such that on $N$ validation examples, the $Np$ highest-scoring are predicted positive. Although we do not take this approach, because at deployment time it is often the case that discrete predictions are required, one could also imagine integrating bias amplification across threshold to have a threshold-agnostic measure of bias amplification, similar to what is proposed by \citet{chen2020threshold}. 

\section{BiasAmp$_{\rightarrow}$ (Directional Bias Amplification)}
\label{sec:our_metrix_exp}
 
Now we present our metric, \biasampours{}, which retains the desirable properties of \biasamptheirs{}, while addressing the
shortcomings noted in Section~\ref{sec:biasamptheirs-shortcomings}. To account for the need to disentangle the two possible directions of bias amplification (shortcoming 3) the metric consists of two values: \biasampoursAT{} corresponds to the amplification of bias resulting from the protected attribute influencing the task prediction, and \biasampoursTA{} corresponds to the amplification of bias resulting from the task influencing the protected attribute prediction. 
Concretely, 
our directional bias amplification metric is:
\begin{align}
  &\text{BiasAmp}_{\rightarrow} = 
    \frac{1}{|\mathcal{A}||\mathcal{T}|}
    \sum_{a\in \mathcal{A},t\in \mathcal{T}}
    y_{at}\Delta_{at}  + (1-y_{at})(-\Delta_{at})
\notag \\
&y_{at} = \mathbbm{1}\left [P(A_a=1,T_t=1) > P(A_a=1)P(T_t=1) \right] 
\notag
\\
&\Delta_{at} = \left\{
                \begin{array}{ll}
                  P(\hat{T}_t=1|A_a=1) - P(T_t=1|A_a=1) & \\ \text{if measuring }A \rightarrow T& \\
                  P(\hat{A}_a=1|T_t=1) - P(A_a=1|T_t=1) & \\\text{if measuring }T \rightarrow A&
                \end{array}
              \right.
\label{eqn:oursfull} 
\end{align}


The first line
generalizes \biasamptheirs{} to include all attributes $A_a$ and measure the amplification of their positive or negative correlations with task $T_t$ (shortcoming 1). The new $y_{at}$ identifies the direction of correlation of $A_a$ with $T_t$, properly accounting for base rates (shortcoming 2). Finally, $\Delta_{at}$ decouples the two possible directions of bias amplification (shortcoming 3). Since values may be negative, reporting the aggregated bias amplification value could obscure attribute-task pairs that exhibit strong bias amplification; thus, disaggregated results per pair can be returned.






\begin{table}
\includegraphics[width=.95\linewidth]{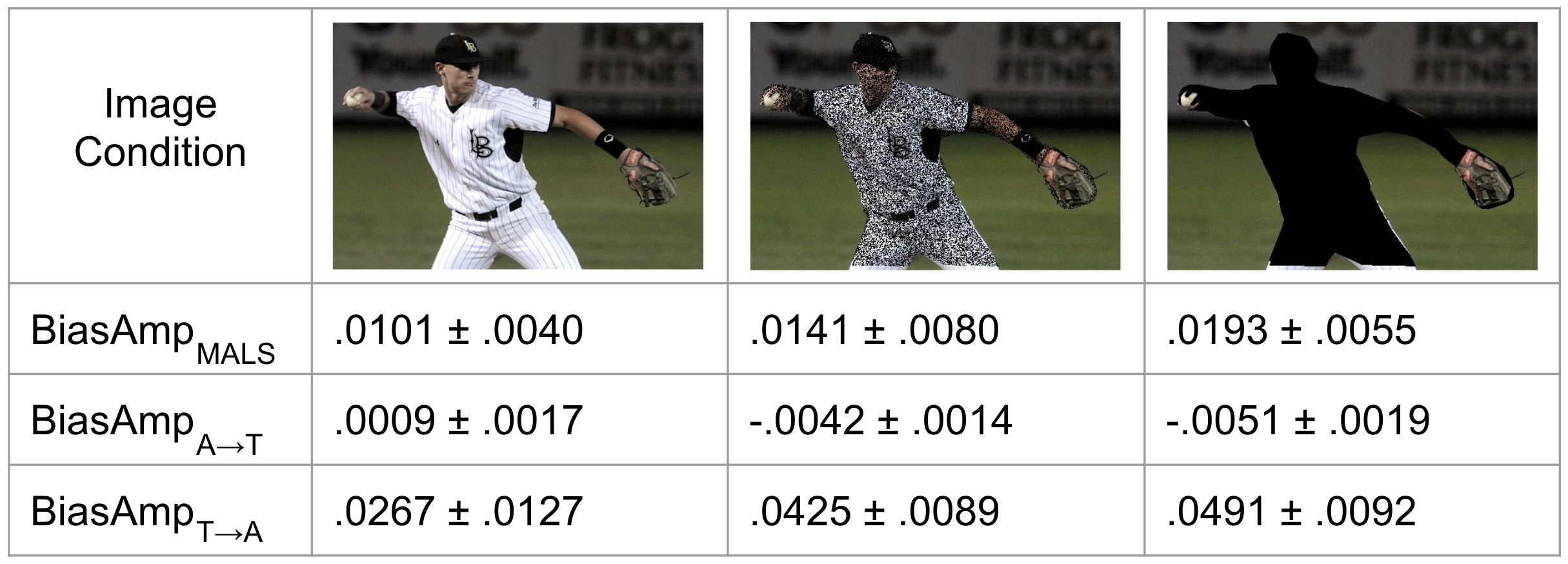}
\caption{Bias amplification, as measured on the test set, changes across three image conditions: original, noisy person mask, full person mask. \biasamptheirs{} misleadingly makes it appear as if bias amplification \emph{increases} as the gender cues are removed. In reality, A$\rightarrow$T decreases with less visual attribute cues to bias the task prediction, while it is T$\rightarrow$A that increases as the model relies on visual cues from the task to predict the attribute.}
  \label{fig:coco_mask}
\end{table}

\subsection{Experimental analysis}
We verify that our metric successfully resolves the empirical inconsistencies of Sec.~\ref{sec:biasamptheirs-shortcomings}. As expected, \biasampoursAT{} is positive at .1778 in shortcoming 1 and .3333 in 2; \biasampoursTA{} is 0 in both. 
We further introduce a scenario for empirically validating the decoupling aspect of our metric. We use a baseline amplification removal idea of applying segmentation masks (noisy or full) over the people in an image to mitigate bias stemming from human attributes~\citep{wang2019balanced}. We train on the COCO classification task, with the same 66 objects from~\citet{zhao2017menshop}, a VGG16~\citep{simonyan2014vgg} model pretrained on ImageNet~\citep{imagenet} to predict objects and gender, with a Binary Cross Entropy Loss over all outputs, and measure \biasampoursTA{} and \biasampoursAT{}; we report 95\% confidence intervals for 5 runs of each scenario. In Tbl.~\ref{fig:coco_mask} we see that, misleadingly, \biasamptheirs{} reports increased amplification as gender cues are removed. However what is actually happening is, as expected, that as less of the person is visible, A$\rightarrow$T decreases because there are less human attribute visual cues to bias the task prediction. It is T$\rightarrow$A that increases because the model must lean into task biases to predict the person's attribute. However, we can also see from the overlapping confidence intervals that the difference between noisy and full masks does not
appear to be particularly robust; we continue a discussion of this phenomenon in Sec.~\ref{sec:consistency}.\footnote{This simultaneously serves as inspiration for an intervention approach to mitigating bias amplification. In
Appendix A.4
we provide a more granular analysis of this experiment, and how it can help to inform mitigation. Further mitigation techniques are outside of our scope, but we look to works like~\citet{singh2020context, wang2019balanced, agarwal2020spurious}. }

\section{Analysis and Discussion}
We now discuss some of the normative issues surrounding bias amplification: in Sec.~\ref{sec:ta_bias} with the existence of T$\rightarrow$A bias amplification, which implies the prediction of sensitive attributes; in Sec.~\ref{sec:consistency} about the need for confidence intervals to make robust conclusions; and in Sec.~\ref{sec:usecase} about scenarios in which the original formulation of bias amplification as a desire to match base correlations may not be the intention.

\subsection{Considerations of T $\rightarrow$ A Bias Amplification}
\label{sec:ta_bias}

\begin{figure}[t]
\centering
\includegraphics[width=1.\linewidth]{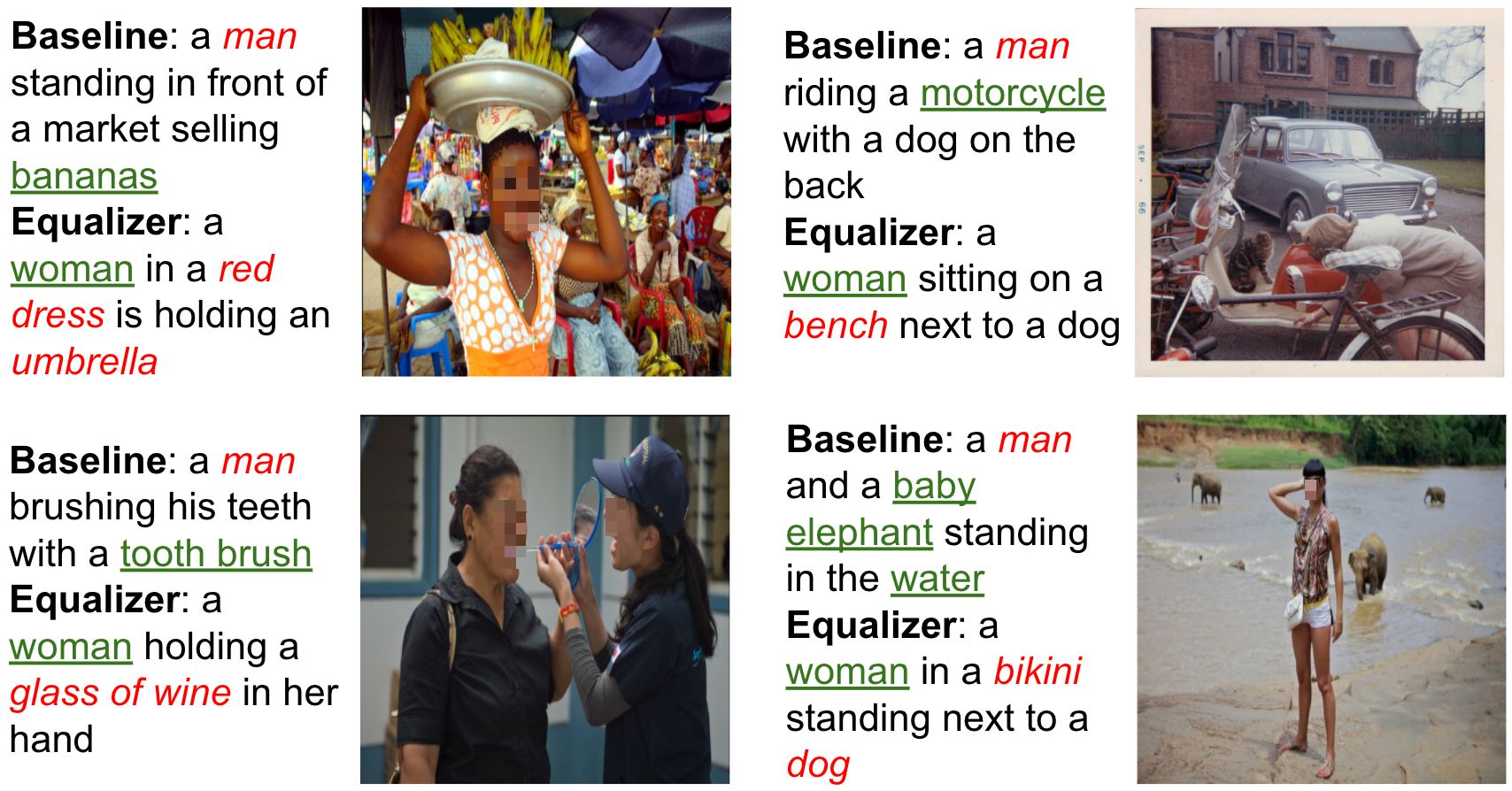} 
\caption{\label{fig:caption}Illustrative captions from the Equalizer model~\citep{hendricks2018snowboard}, which in these captions decreases $T\rightarrow A$ bias amplification from the Baseline, but inadvertently increases $A\rightarrow T$. Green underlined words are correct, and red italicized words are incorrect. In the images, the Equalizer improves on the Baseline for the gendered word, but introduces biased errors in the captions.}
\end{figure}

If we think more deeply about these bias amplifications, we might come to a normative conclusion that $T\rightarrow A$, which measures sensitive attribute predictions conditioned on the tasks, should not exist in the first place. There are very few situations in which predicting sensitive attributes makes sense~\citep{scheuerman2020identity, larson2017gender},
so we should carefully consider if this is strictly necessary for target applications. For the image domains discussed, by simply removing the notion of predicting gender, we trivially remove all $T\rightarrow A$ bias amplification. In a similar vein, there has been great work done on reducing gender bias in image captions~\citep{hendricks2018snowboard, tang2020caption}, but it is often focused on targeting $T\rightarrow A$ rather than $A\rightarrow T$ amplification. When disentangling the directions of bias, we find that the Equalizer model~\citep{hendricks2018snowboard}, which was trained with the intention of increasing the quality of gender-specific words in captions, inadvertently increases $A\rightarrow T$ bias amplification for certain tasks. We treat gender as the attribute and the objects as different tasks. In Fig.~\ref{fig:caption} we see examples where the content of the Equalizer's caption exhibits bias coming from the person's attribute. 
Even though the Equalizer model reduces $T\rightarrow A$ bias amplification in these images, it inadvertently increases $A\rightarrow T$. It is important to disentangle the two directions of bias and notice that while one direction is becoming more fair, another is actually becoming more biased. Although this may not always be the case, depending on the downstream application~\cite{bennett2021accessibility}, perhaps we could consider simply replacing all instances of gendered words like ``man" and ``woman" in the captions  with ``person" to trivially eliminate $T\rightarrow A$, and focus on $A\rightarrow T$ bias amplification.
Specifically when gender is the sensitive attribute, ~\citet{keyes2018agr} thoroughly explains how we should carefully think about why we might implement Automatic Gender Recognition (AGR), and avoid doing so. 

On the other hand, sensitive attribute labels, ideally from self-disclosure, can be very useful. For example, these labels are necessary to measure $A\rightarrow T$ amplification, which is important to discover, as we do not want our prediction task to be biased for or against people with certain attributes.

\subsection{Variance in Estimator Bias}
\label{sec:consistency}
Evaluation metrics, ours included, are specific to each model on each dataset. Under common loss functions such as cross entropy loss, some evaluation metrics like average precision are not very sensitive to random seed. However, bias amplification, along with other fairness metrics like FPR difference, often fluctuates greatly across runs. Because the loss functions that machine learning practitioners tend to default to using are proxies for accuracy, it makes sense that various local minima, while equal in accuracy, are not necessarily equal for other measurements. The phenomena of differences between equally predictive models has been termed the Rashomon Effect~\citep{breiman2001cultures}, or predictive multiplicity~\citep{marx2020multiplicity}.


Thus, like previous work~\citep{fisher2019wrong}, we urge transparency, and advocate for the inclusion of confidence intervals. To illustrate the need for this, we look at the facial image domain of CelebA~\citep{liu2015faceattributes}, defining two different scenarios of the classification of \texttt{big nose} or \texttt{young} as our task, and treating the gender labels as our attribute.
Note that we do not classify gender, for reasons raised in Sec.~\ref{sec:ta_bias}, so we only measure $A \rightarrow T$ amplification. For these tasks, women are correlated with no big nose and being young, and men with big nose and not being young. We examine two different scenarios, one where our independent variable is model architecture, and another where it is the ratio between number of images of the majority groups (e.g., young women and not young men) and minority groups (e.g., not young women and young men). By looking at the confidence intervals, we can determine which condition allows us to draw reliable conclusions about the impact of that variable on bias amplification.

For model architecture, we train 3 models pretrained on ImageNet~\citep{imagenet} across 5 runs: ResNet18~\citep{he16resnet}, AlexNet~\citep{alexnet}, and VGG16~\citep{simonyan2014vgg}. Training details
are in 
Appendix A.2.
In Fig.~\ref{fig:models} we see from the confidence intervals that while model architecture does not result in differing enough of bias amplification to conclude anything about the relative fairness of these models, across-ratio differences are significant enough to draw conclusions about the impact of this ratio on bias amplification. We encourage researchers to include confidence intervals so that findings are more robust to random fluctuations. Concurrent work covers this multiplicity phenomenon in detail~\citep{damour2020underspecification}, and calls for more application-specific specifications that would constrain the model space. However, that may not always be feasible, so for now our proposal of error bars is more general and immediately implementable. In a survey of recently published fairness papers from prominent machine learning conferences, we found that 25 of 48 (52\%) reported results of a fairness metric without error bars (details in
Appendix A.2.
Even if the model itself is deterministic, error bars could be generated through bootstrapping~\citep{efron1992bootstrap} to account for the fact that the test set itself is but a sample of the population, or varying the train-test splits~\citep{friedler2019comparative}.

\begin{figure*}[t]
\includegraphics[width=0.94\linewidth]{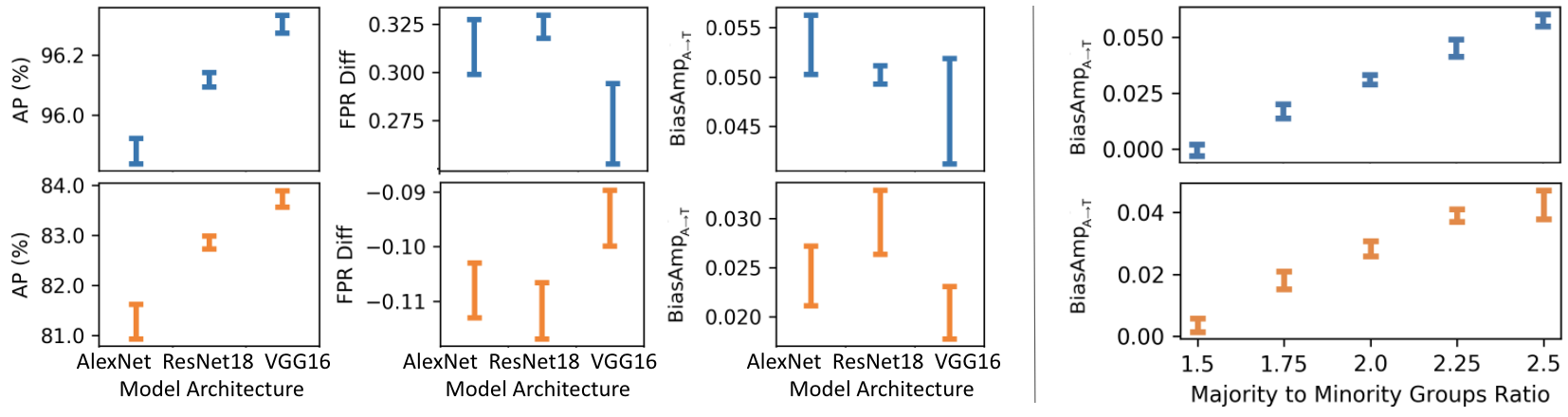} 
\caption{\label{fig:models}We investigate the consistency of various metrics by looking at 95\% confidence intervals as we manipulate two independent variables: model architecture (left three graphs), and majority to minority groups ratio (right graph). The top row (blue) is for the attribute of \texttt{big nose}, and bottom row (orange) is for \texttt{young}. For model architecture, across 5 runs, the accuracy measure of average precision retains a consistent ranking across models, but two different fairness measures (FPR difference and $A\rightarrow T$ bias amplification) have overlapping intervals. This does not allow us to draw conclusions about the differing fairness of these models. However, across-ratio differences in bias amplification are significant enough to allow us to draw conclusions about the differing levels of fairness.} 
\end{figure*}

\subsection{Limitations of Bias Amplification}
\label{sec:usecase}


An implicit assumption that motivates bias amplification metrics, including ours, is that the ground truth exists and is known. Further, a perfectly accurate model can be considered perfectly fair, despite the presence of task-attribute correlations in the training data. This allows us to treat the disparity between the correlations in the input vs correlations in the output as a fairness measure. 

It follows that bias amplification would {\em not} be a good way to measure fairness when the ground truth is either unknown or does not correspond to desired classification. In this section, we discuss two types of applications where bias amplification should not necessarily be used out-of-the-box as a fairness metric. 





\smallsec{Sentence completion: no ground truth}
\label{sec:nlp}
Consider the fill-in-the-blank NLP task, where there is no ground truth for how to fill in a sentence. Given ``The [blank] went on a walk", a variety of words could be suitable. Therefore, to measure bias amplification in this setting, we need to subjectively set the base correlations, i.e., $P(T_t=1|A_a=1), P(A_a=1|T_t=1)$. 

To see the effect of adjusting base correlations, we test the bias amplification between occupations and gender pronouns, conditioning on the pronoun and filling in the occupation and vice versa. In Tbl.~\ref{tbl:nlp}, we report our measured bias amplification results on the FitBERT (Fill in the blanks BERT)~\citep{havens2019fitbert, devlin2019bert} model using various sources as our base correlation of bias from which amplification is measured. The same outputs from the model are used for each set of pronouns, and the independent variable we manipulate is the source of base correlations: 1) equality amongst the pronouns, using two pronouns (he/she) 2) equality amongst the pronouns, using three pronouns (he/she/they) 3) co-occurrence counts from English Wikipedia (one of the datasets BERT was trained on), and 4) WinoBias~\citep{zhao2018winobias} with additional information supplemented from the 2016 U.S. Labor Force Statistics data. 
Additional details are in
Appendix A.3.

\begin{figure}[t]
\centering
\includegraphics[width=0.9\linewidth]{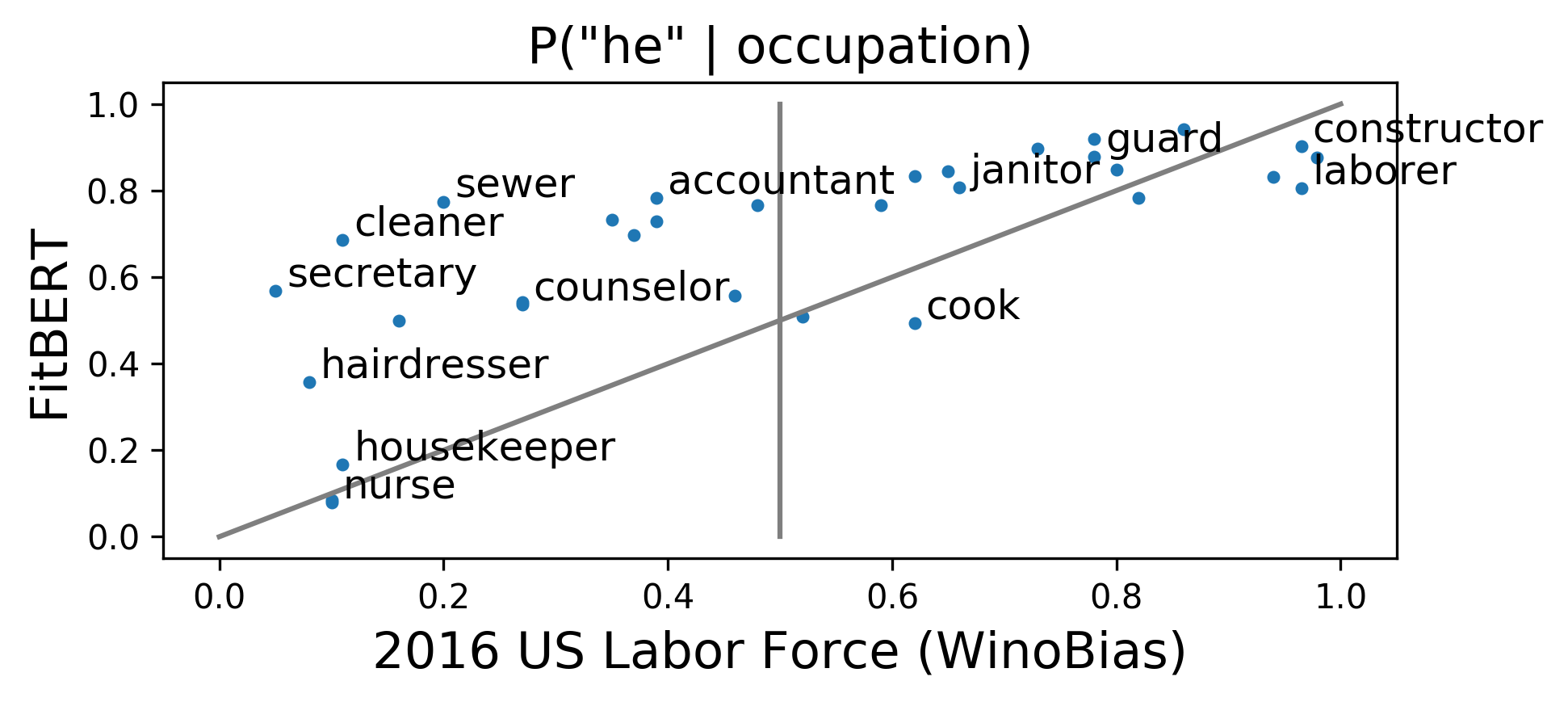} 
\caption{\label{fig:occu_bert}Each point represents an occupation's probability at being associated with the pronoun for a man. FitBERT perpetuates gender-occupation biases seen in the U.S. Labor Force, and additionally over-favors the pronoun for men.}
\end{figure}

We find that relative to U.S. Labor Force data on these particular occupations, FitBERT actually exhibits no bias amplification. Yet it would be simplistic to conclude that FitBERT presents no fairness concerns with respect to gender and occupation. For one, it is evident from Fig.~\ref{fig:occu_bert} that there is an overall bias towards ``he" (this translates to a bias amplification for some occupations and a bias reduction for others; the effects roughly cancel out in our bias amplification metric when aggregated). More importantly, whether U.S. labor statistics are the right source of base correlations depends on the specific application of the model and the cultural context in which it is deployed. This is clear when noticing that the measured \biasampoursTA{} is much stronger when the gender distribution is expected to be uniform, instead of gender-biased Labor statistics.


\begin{table}
\begin{center}
\begin{tabular}{ |>{\centering\arraybackslash}p{2.94cm}|>{\centering\arraybackslash}p{2.1cm}|>{\centering\arraybackslash}p{2.1cm}| } 
 \hline
 Base Correlation Source (\# pronouns) & \biasampoursTA{} & \biasampoursAT{} \\ 
 \hline
 Uniform (2) & .1368 $\pm$ .0226 & .0084 $\pm$ .0054 \\ 
 \hline
 Uniform (3) & .0914 $\pm$ .0151 & .0056 $\pm$ .0036 \\ 
 \hline
 Wikipedia (2) &  .0372 $\pm$ .0307 & -.0002 $\pm$ .0043 \\ 
 \hline
 2016 U.S. Labor Force (WinoBias) (2) & -.1254 $\pm$ .0026 & -.0089 $\pm$ .0054 \\ 
 \hline
\end{tabular}
\end{center}
\caption{\label{tbl:nlp}\biasampours{} for different base correlation sources. The value-laden choice of base correlation source depends on the downstream application.}
\end{table}

\smallsec{Risk prediction: future outcomes unknown}
Next, we examine the criminal risk prediction setting. A common statistical task in this setting is predicting the likelihood a defendant will commit a crime if released pending trial. This setting has two important differences compared to computer vision detection tasks: 1) The training labels typically come from arrest records and suffer from problems like historical and selection bias~\citep{suresh2019framework, olteanu2019social, green2020recidivism}, and 2) the task is to predict future events and thus the outcome is not knowable at prediction time. Further, the risk of recidivism is not a static, immutable trait of a person. Given the input features that are used to represent individuals, one could imagine an individual with a set of features who does recidivate, and one who does not. In contrast, for a task like image classification, two instances with the same pixel values will always have the same labels (if the ground truth labels are accurate).

As a result of these setting differences, risk prediction tools may be considered unfair even if they exhibit no bias amplification. Indeed, one might argue that a model that shows no bias amplification is necessarily unfair as it perpetuates past biases reflected in the training data. Further, modeling risk as immutable misses the opportunity for intervention to change the risk~\cite{barabas2018intervention}. Thus, matching the training correlations should not be the intended goal~\citep{wick2019tradeoff, hebertjohnson2018multicalibration}.

\begin{figure}[t]
\centering
\includegraphics[width=0.9\linewidth]{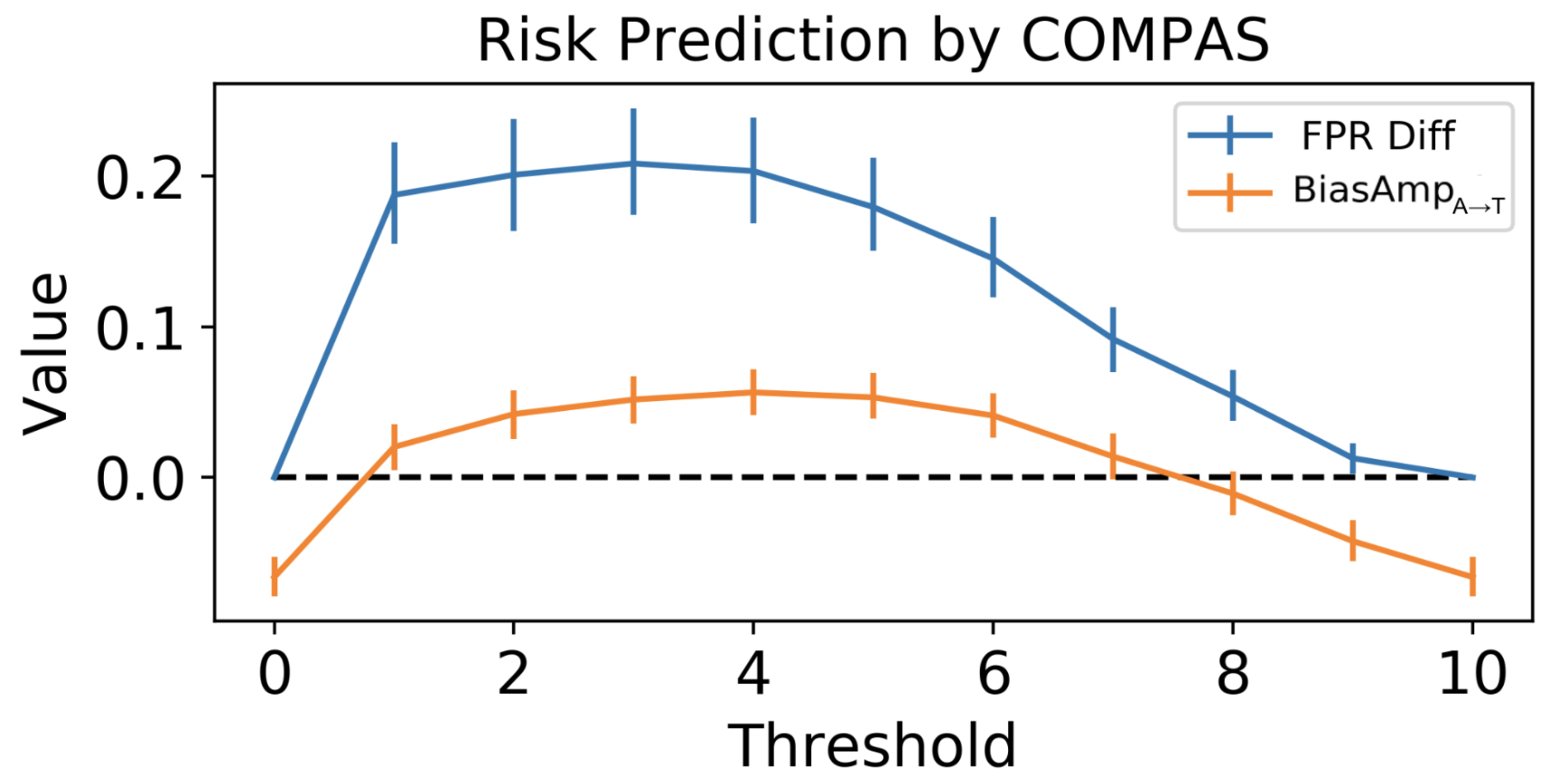} 
\caption{\label{fig:compas}COMPAS risk predictions exhibit FPR disparities, but little bias amplification. Bias amplification measures only whether the model matches the (biased) training data, not the bias of the overall system.}
\end{figure}

To make this more concrete, in Fig.~\ref{fig:compas} we show the metrics
of \biasampoursAT{} and False Positive Rate (FPR) disparity measured on COMPAS predictions~\citep{angwin2016compas}, only looking at two racial groups, for various values of the risk threshold. A false positive occurs when  a defendant classified as high risk but does not recidivate; FPR disparity has been interpreted as measuring how unequally different groups suffer the costs of the model's errors \cite{hardt2016equalopp}. The figure shows that bias amplification is close to 0 for almost all thresholds. This is no surprise since the model was designed to be calibrated by group~\cite{flores2016compas}. However, for all realistic values of the threshold, there is a large FPR disparity. Thus, risk prediction is a setting where a lack of bias amplification should not be used to conclude that a model is fair.

Like any fairness metric, ours captures only one perspective, which is that of not amplifying already present biases. It does not require a correction for these biases. Settings that bias amplification are more suited for include those with a known truth in the labels, where matching them would desired. For example, applicable contexts include certain social media bot detection tasks where the sensitive attribute is the region of origin, as bot detection methods may be biased against names from certain areas. More broadly, it is crucial that we pick fairness metrics thoughtfully when deciding how to evaluate a model. 


\section{Conclusion}
In this paper, we take a deep dive into the measure of bias amplification. We introduce a new metric, \biasampours{},
that disentangles the directions of bias to provide more actionable insights when diagnosing models. Additionally, we analyze and discuss normative considerations
to encourage exercising care when determining which fairness metrics are applicable, and what assumptions they are encoding. The mission of this paper is not to tout bias amplification as the optimal fairness metric, but rather to give a comprehensive and critical study as to how it should be measured.


\section*{Acknowledgements}
This material is based upon work supported by the National Science Foundation under Grant No. 1763642. We thank Sunnie S. Y. Kim, Karthik Narasimhan, Vikram Ramaswamy, Brandon Stewart, and Felix Yu for feedback. We especially thank Arvind Narayanan for significant comments and advice.
We also thank the authors of Men Also Like Shopping~\citep{zhao2017menshop} and Women Also Snowboard~\citep{hendricks2018snowboard} for uploading their model outputs and code online in a way that made it easily reproducible, and for being prompt and helpful in response to clarifications.


\bibliography{bibs}
\bibliographystyle{icml2021}





\onecolumn
\appendix
\section{Appendix}
\subsection{Additional Metric Details}
\label{app:metricdetails}


We provide additional details here about \biasampours{}, as defined in Sec. 4.

In practice the indicator variable, $y_{at}$, is computed over the statistics of the training set, whereas everything else is computed over the test set. The reason behind this is that the direction of bias is determined by the existing biases in the training set. 

Comparisons of the values outputted by \biasampours{} should only be done relatively. In particular, within one of the directions at a time, either $A\rightarrow T$ or $T\rightarrow A$, on one dataset. Comparing $A\rightarrow T$ to $T\rightarrow A$ directly is not a signal as to which direction of amplification is stronger.

\subsection{Details and Experiment from Variance in Estimator Bias}
\label{app:multiplicity_details}

For the models we trained in Sec. 5.2, we performed hyperparameter tuning on the validation set, and ended up using the following: ResNet18 had a learning rate of .0001, AlexNet of .0003, and VGG16 of .00014. All models were trained with stochastic gradient descent, a batch size of 64, and 10 epochs. We use the given train-validation-test split from the CelebA dataset.

Our method for surveying prominent fairness papers is as follows: on Google Scholar we performed a search for papers containing the keywords of ``fair", ``fairness", or ``bias" from the year 2015 onwards, sorted by relevance. We did this for the three conferences of 1) Conference on Neural Information Processing Systems (NeurIPS), 2) International Conference on Machine Learning (ICML), and 3) ACM Conference on Fairness, Accountability, and Transparency (FAccT). We picked these conferences because of their high reputability as machine learning conferences, and thus would serve as a good upper bound for reporting error bars on fairness evaluation metrics. We also looked at the International Conference on Learning Representations (ICLR), but the Google Scholar search turned up very few papers on fairness. From the three conferences we ended up querying, we took the first 25 papers from each, pruning those that were either: 1) not related to fairness, or 2) not containing fairness metrics for which it error bars could be relevant (e.g., theoretical or philosophical papers). Among the 48 papers that were left of the 75, if there was at least one graph or table containing a fairness metric that did not appear to be fully deterministic, and no error bars were included (even if the number reported was a mean across multiple runs), this was marked to be a ``non-error-bar" paper, of which 25 of the 48 papers looked into met this criteria.

\subsection{Details on Measuring Bias Amplification in FitBERT}
\label{app:nlp}
Here we provide additional details behind the numbers presented in Tbl. 2 in Sec. 5.3.

As noted, and done, by~\cite{liang2020big}, a large and diverse corpus of sentences is needed to sample from the large variety of contexts. However, that is out of scope for this work, where we run FitBERT on 20 sentence templates of the form ``[1) he/she/(they)] [2) is/was] a(n) [3) adjective] [4) occupation]". By varying 2) and using the top 10 most frequent adjectives from a list of adjectives~\citep{english_adj} that appear in the English Wikipedia dataset (one of the datasets BERT was trained on) that would be applicable as a descriptor for an occupation (pruning adjectives like e.g., ``which", ``left") for 3), we end up with 20 template sentences. We then alternate conditioning on 1) (to calculate A$\rightarrow$T) and 4) (to calculate T$\rightarrow$A). The 10 adjectives we ended up wtih are: new, known, single, large, small, major, French, old, short, good. We use the output probabilities rather than discrete predictions in calculating $P(\hat{A}_a=1|T_t=1)$ and $P(\hat{T}_t=1|A_a=1)$ because there is no ``right" answer in sentence completion, in contrast to object prediction, and so we want the output distribution.

When calculating the amount of bias amplification when the base rates are equal, we picked the direction of bias based on that provided by the WinoBias dataset. In practice, this can be thought of as setting the base correlation, $P(A_a=1|T_t=1)$ for a men-biased job like ``cook" to be $.5+\epsilon$ for ``he" and $.5-\epsilon$ for ``she" when there are two pronouns, and $.33+\epsilon$ for ``he" and $.33-\epsilon$ for ``she" and ``they", where in practice we used $\epsilon=1\mathrm{e-}7$. This ensures that the indicator variable, $y_{at}$ from Eq. 2, is set in the direction fo the gender bias, but the magnitudes of $\Delta_{at}$ are not affected to a significant degree.

To generate a rough approximation of what training correlation rates could look like in this domain, we look to one of the datasets that BERT was trained on, the Wikipedia dataset. We do so by simply counting the cooccurrences of all the occupations along with gendered words such as ``man", ``he", ``him", etc. There are flaws with this approach because in a sentence like ``She went to see the doctor.", the pronoun is in fact not referring to the gender of the person with the occupation. However, we leave a more accurate measurement of this to future work, as our aim for showing these results was more for demonstrative purposes illustrating the manipulation of the correlation rate, rather than in rigorously measuring the training correlation rate.

We use 32 rather than 40 occupations in WinoBias~\citep{zhao2018winobias}, because when we went to the 2016 U.S. Labor Force Statistics data~\citep{labor2016} to collect the actual numbers of each gender and occupation in order to be able to calculate $P(T_t=1|A_a=1)$, since WinoBias only had $P(A_a=1|T_t=1)$, we found 8 occupations to be too ambiguous to be able to determine the actual numbers. For example, for ``attendant", there were many different attendant jobs listed, such as ``flight attendants" and ``parking lot attendant", so we opted rather to drop these jobs from the list of 40. The 8 from the original WinoBias dataset that we ignored are: supervisor, manager, mechanician, CEO, teacher, assistant, clerk, and attendant. The first four are biased towards men, and the latter four towards women, so that we did not skew the distribution of jobs biased towards each gender.

\subsection{COCO Masking Experiment Broken Down by Object}
\label{app:brokendown_mask}
In Table~\ref{fig:coco_mask} of Sec.~\ref{sec:our_metrix_exp} we perform an experiment whereby we measure the bias amplification on COCO object detection based on the amount of masking we apply to the people in the images. We find that  \biasampoursAT decreases when we apply masking, but \biasampoursTA increases when we do so. To better inform mitigation techniques, it is oftentimes helpful to take a more granular look at which objects are actually amplifying the bias. In Table~\ref{tbl:breakdown} we provide such a granular breakdown. If our goal is to target \biasampoursAT, we might note that objects like \texttt{tv} show decreasing bias amplification when the person is masked, while \texttt{dining table} stays relatively stagnant.

  \begin{table}[t]
    \caption{A breakdown of \biasampoursAT and \biasampoursTA by object for the masking experiment done on COCO in Table~\ref{fig:coco_mask}.}
  \label{tbl:breakdown}
  \centering
   \begin{tabular}{p{.65in}|>{\centering\arraybackslash}p{.78in}|>{\centering\arraybackslash}p{.78in}|>{\centering\arraybackslash}p{.78in}|>{\centering\arraybackslash}p{.78in}|>{\centering\arraybackslash}p{.78in}|>{\centering\arraybackslash}p{.78in}}
    \toprule
    \multirow{2}{*}{Object} & \multicolumn{2}{c|}{Original} & \multicolumn{2}{c|}{Noisy Person Mask}& \multicolumn{2}{c}{Full Person Mask}  \\
    \cline{2-7}
    & $A\rightarrow T$ & $T\rightarrow A$& $A\rightarrow T$ & $T\rightarrow A$& $A\rightarrow T$ & $T\rightarrow A$ \\
    \hline
teddy bear &  $-0.13 \pm 0.04$ &    $1.23 \pm 0.32$ &   $0.13 \pm 0.04$ &    $2.27 \pm 0.34$ &  $-0.09 \pm 0.04$ &    $1.93 \pm 0.43$ \\
        handbag &   $0.44 \pm 0.14$ &    $7.88 \pm 0.67$ &    $0.4 \pm 0.17$ &    $5.62 \pm 2.13$ &   $0.24 \pm 0.13$ &    $2.96 \pm 3.03$ \\
           fork &   $0.62 \pm 0.22$ &    $7.67 \pm 1.31$ &   $0.63 \pm 0.18$ &    $7.45 \pm 1.65$ &   $0.76 \pm 0.24$ &    $6.22 \pm 0.62$ \\
           cake &  $-0.29 \pm 0.06$ &     $5.7 \pm 0.59$ &  $-0.09 \pm 0.04$ &     $5.2 \pm 0.59$ &  $-0.16 \pm 0.06$ &      $3.3 \pm 1.7$ \\
            bed &   $0.01 \pm 0.04$ &    $1.33 \pm 1.43$ &   $0.06 \pm 0.08$ &    $5.33 \pm 1.43$ &  $-0.09 \pm 0.03$ &     $6.67 \pm 0.0$ \\
       umbrella &  $-0.05 \pm 0.07$ &    $9.52 \pm 3.92$ &   $0.08 \pm 0.06$ &   $13.33 \pm 2.64$ &  $-0.04 \pm 0.11$ &   $11.24 \pm 2.76$ \\
          spoon &   $0.06 \pm 0.06$ &   $-2.91 \pm 2.78$ &   $0.03 \pm 0.06$ &  $-10.91 \pm 4.83$ &   $0.04 \pm 0.03$ &  $-15.27 \pm 4.11$ \\
        giraffe &   $0.21 \pm 0.13$ &    $5.32 \pm 1.51$ &    $0.2 \pm 0.05$ &    $4.05 \pm 2.46$ &   $0.02 \pm 0.04$ &    $4.95 \pm 1.34$ \\
           bowl &   $0.28 \pm 0.04$ &     $2.2 \pm 1.02$ &   $0.06 \pm 0.07$ &     $4.8 \pm 2.56$ &   $0.18 \pm 0.12$ &     $7.4 \pm 2.26$ \\
          knife &  $-0.22 \pm 0.12$ &  $-11.74 \pm 2.39$ &  $-0.35 \pm 0.12$ &  $-10.43 \pm 3.15$ &  $-0.31 \pm 0.05$ &   $-8.84 \pm 2.18$ \\
     wine glass &  $-0.41 \pm 0.14$ &   $-5.24 \pm 0.83$ &  $-0.62 \pm 0.09$ &   $-7.14 \pm 3.49$ &  $-0.69 \pm 0.07$ &  $-10.95 \pm 3.64$ \\
   dining table &  $-0.75 \pm 0.14$ &    $4.53 \pm 1.02$ &  $-0.76 \pm 0.14$ &    $3.18 \pm 2.12$ &  $-0.74 \pm 0.17$ &    $4.58 \pm 1.38$ \\
            cat &   $0.07 \pm 0.04$ &    $2.41 \pm 2.05$ &   $0.19 \pm 0.05$ &    $10.0 \pm 2.42$ &   $0.17 \pm 0.02$ &    $20.0 \pm 1.81$ \\
           sink &   $0.18 \pm 0.15$ &    $-4.29 \pm 2.4$ &   $0.11 \pm 0.07$ &    $-5.03 \pm 2.6$ &   $-0.12 \pm 0.1$ &    $-3.19 \pm 1.5$ \\
            cup &   $-0.3 \pm 0.09$ &  $-12.36 \pm 3.42$ &  $-0.15 \pm 0.08$ &   $-12.0 \pm 1.32$ &  $-0.12 \pm 0.03$ &  $-14.42 \pm 2.94$ \\
   potted plant &   $0.21 \pm 0.18$ &     $6.36 \pm 5.1$ &   $0.34 \pm 0.07$ &     $0.0 \pm 2.82$ &   $0.32 \pm 0.08$ &   $-4.55 \pm 2.52$ \\
   refrigerator &  $-0.06 \pm 0.03$ &     $9.6 \pm 1.36$ &  $-0.07 \pm 0.06$ &    $7.47 \pm 1.75$ &   $0.01 \pm 0.03$ &    $12.0 \pm 4.25$ \\
      microwave &  $-0.01 \pm 0.02$ &     $-3.5 \pm 5.3$ &  $-0.01 \pm 0.03$ &   $13.0 \pm 13.32$ &  $-0.03 \pm 0.03$ &     $6.0 \pm 9.05$ \\
          couch &  $-1.35 \pm 0.16$ &   $-0.25 \pm 1.24$ &  $-0.94 \pm 0.23$ &    $1.62 \pm 1.06$ &  $-1.21 \pm 0.16$ &     $0.15 \pm 0.8$ \\
           oven &   $0.07 \pm 0.09$ &    $7.67 \pm 1.73$ &  $-0.33 \pm 0.12$ &    $10.67 \pm 3.0$ &   $0.03 \pm 0.13$ &   $13.12 \pm 1.49$ \\
       sandwich &  $-0.98 \pm 0.15$ &   $-8.93 \pm 0.92$ &   $-2.6 \pm 0.18$ &  $-15.23 \pm 2.78$ &   $-2.46 \pm 0.4$ &  $-15.67 \pm 1.74$ \\
           book &  $-0.43 \pm 0.08$ &   $-3.07 \pm 0.57$ &  $-0.48 \pm 0.13$ &   $-3.34 \pm 1.24$ &  $-0.85 \pm 0.11$ &    $-3.18 \pm 2.0$ \\
         bottle &   $0.05 \pm 0.11$ &   $-7.73 \pm 1.54$ &  $-0.13 \pm 0.14$ &   $-8.33 \pm 3.87$ &   $0.06 \pm 0.06$ &  $-11.52 \pm 1.65$ \\
     cell phone &  $-0.09 \pm 0.13$ &     $3.6 \pm 1.92$ &   $0.05 \pm 0.15$ &   $13.72 \pm 0.89$ &   $-0.1 \pm 0.12$ &   $18.72 \pm 2.36$ \\
          pizza &   $-0.19 \pm 0.1$ &    $6.85 \pm 1.81$ &  $-0.09 \pm 0.03$ &   $15.17 \pm 2.41$ &  $-0.38 \pm 0.12$ &     $9.3 \pm 1.56$ \\
         banana &   $0.35 \pm 0.08$ &    $4.42 \pm 0.66$ &   $0.56 \pm 0.09$ &     $6.1 \pm 1.06$ &    $0.1 \pm 0.19$ &    $5.19 \pm 0.72$ \\
     toothbrush &  $-0.47 \pm 0.11$ &   $-2.42 \pm 2.63$ &  $-0.55 \pm 0.13$ &   $-4.32 \pm 5.14$ &   $-0.5 \pm 0.12$ &  $-11.85 \pm 4.43$ \\
  tennis racket &  $-0.09 \pm 0.08$ &    $9.22 \pm 3.05$ &    $0.0 \pm 0.09$ &   $14.75 \pm 2.74$ &   $0.09 \pm 0.12$ &   $14.75 \pm 1.38$ \\
          chair &  $-0.31 \pm 0.11$ &    $1.87 \pm 0.22$ &  $-0.31 \pm 0.06$ &    $3.85 \pm 0.37$ &  $-0.31 \pm 0.05$ &     $4.69 \pm 0.0$ \\
            dog &   $0.14 \pm 0.04$ &    $0.43 \pm 0.19$ &    $0.3 \pm 0.07$ &     $1.6 \pm 0.29$ &    $0.3 \pm 0.04$ &     $1.7 \pm 0.35$ \\
          donut &   $-0.3 \pm 0.08$ &    $-1.4 \pm 0.29$ &  $-0.39 \pm 0.15$ &    $-0.0 \pm 0.59$ &  $-0.41 \pm 0.15$ &    $0.09 \pm 0.37$ \\
       suitcase &  $-0.43 \pm 0.08$ &    $1.96 \pm 1.12$ &  $-0.09 \pm 0.01$ &     $7.3 \pm 0.65$ &  $-0.11 \pm 0.14$ &     $8.43 \pm 1.4$ \\
         laptop &   $0.27 \pm 0.07$ &    $4.58 \pm 3.57$ &   $0.06 \pm 0.04$ &    $7.67 \pm 6.33$ &    $0.1 \pm 0.06$ &   $17.39 \pm 5.52$ \\
        hot dog &   $1.48 \pm 0.12$ &    $7.63 \pm 2.72$ &   $1.51 \pm 0.07$ &    $9.37 \pm 1.75$ &   $1.86 \pm 0.14$ &    $9.16 \pm 3.03$ \\
         remote &   $0.33 \pm 0.02$ &    $9.14 \pm 2.73$ &   $0.15 \pm 0.09$ &   $11.03 \pm 2.59$ &   $0.15 \pm 0.07$ &   $18.62 \pm 3.12$ \\
          clock &   $0.77 \pm 0.16$ &    $4.48 \pm 3.05$ &   $0.62 \pm 0.08$ &    $9.61 \pm 3.02$ &   $0.88 \pm 0.11$ &   $11.44 \pm 3.78$ \\
          bench &  $-0.02 \pm 0.05$ &   $11.49 \pm 3.61$ &  $-0.06 \pm 0.06$ &   $14.68 \pm 4.73$ &  $-0.04 \pm 0.06$ &    $16.6 \pm 2.61$ \\
             tv &   $0.35 \pm 0.09$ &    $5.16 \pm 4.59$ &   $0.27 \pm 0.12$ &   $14.19 \pm 2.58$ &   $0.21 \pm 0.09$ &   $18.06 \pm 3.03$ \\
          mouse &  $-0.22 \pm 0.06$ &    $0.95 \pm 5.76$ &  $-0.26 \pm 0.05$ &    $4.57 \pm 4.99$ &  $-0.18 \pm 0.06$ &    $5.14 \pm 4.11$ \\
          horse &   $0.07 \pm 0.03$ &    $8.13 \pm 5.17$ &   $0.11 \pm 0.07$ &   $13.63 \pm 1.96$ &   $0.16 \pm 0.04$ &   $16.59 \pm 4.92$ \\
   fire hydrant &  $-0.21 \pm 0.07$ &    $4.71 \pm 2.98$ &  $-0.15 \pm 0.07$ &    $1.76 \pm 4.64$ &  $-0.18 \pm 0.06$ &     $5.0 \pm 6.19$ \\
       keyboard &   $0.01 \pm 0.08$ &     $1.64 \pm 3.4$ &  $-0.08 \pm 0.08$ &   $17.38 \pm 4.51$ &   $0.02 \pm 0.08$ &   $31.15 \pm 6.74$ \\
            bus &   $0.02 \pm 0.04$ &  $-11.33 \pm 2.15$ &  $-0.17 \pm 0.07$ &    $-9.0 \pm 3.53$ &   $-0.1 \pm 0.04$ &     $3.0 \pm 6.62$ \\
         toilet &   $0.26 \pm 0.06$ &    $7.65 \pm 4.07$ &    $0.2 \pm 0.09$ &   $12.17 \pm 4.15$ &   $0.19 \pm 0.06$ &   $17.22 \pm 5.67$ \\
         person &   $-0.04 \pm 0.1$ &   $-1.47 \pm 4.54$ &    $0.1 \pm 0.07$ &   $-4.12 \pm 5.25$ &   $0.03 \pm 0.08$ &   $-3.53 \pm 4.65$ \\
  traffic light &   $-0.2 \pm 0.03$ &    $4.44 \pm 2.48$ &  $-0.27 \pm 0.06$ &    $6.06 \pm 1.86$ &  $-0.27 \pm 0.11$ &   $11.52 \pm 2.28$ \\
    sports ball &   $-1.44 \pm 0.2$ &    $3.41 \pm 1.56$ &  $-0.95 \pm 0.27$ &    $3.86 \pm 1.83$ &  $-1.27 \pm 0.32$ &    $4.95 \pm 1.97$ \\
        bicycle &  $-0.23 \pm 0.07$ &   $13.58 \pm 2.02$ &   $0.01 \pm 0.15$ &   $13.09 \pm 3.98$ &  $-0.21 \pm 0.08$ &   $12.72 \pm 1.83$ \\
            car &     $0.2 \pm 0.2$ &     $3.7 \pm 1.49$ &    $0.57 \pm 0.1$ &    $-0.74 \pm 2.2$ &    $0.32 \pm 0.1$ &    $0.37 \pm 2.66$ \\
       backpack &   $0.01 \pm 0.03$ &    $7.57 \pm 1.63$ &    $0.0 \pm 0.06$ &   $17.84 \pm 3.09$ &  $-0.05 \pm 0.04$ &   $18.38 \pm 2.53$

  \end{tabular}
  \end{table}
  
  \setcounter{table}{2}
  \begin{table}[t]
    \caption{\textit{Continued}}
  \label{tbl:breakdown}
  \centering
   \begin{tabular}{p{.65in}|>{\centering\arraybackslash}p{.78in}|>{\centering\arraybackslash}p{.78in}|>{\centering\arraybackslash}p{.78in}|>{\centering\arraybackslash}p{.78in}|>{\centering\arraybackslash}p{.78in}|>{\centering\arraybackslash}p{.78in}}
    \toprule
    \multirow{2}{*}{Object} & \multicolumn{2}{c|}{Original} & \multicolumn{2}{c|}{Noisy Person Mask}& \multicolumn{2}{c}{Full Person Mask}  \\
    \cline{2-7}
    & $A\rightarrow T$ & $T\rightarrow A$& $A\rightarrow T$ & $T\rightarrow A$& $A\rightarrow T$ & $T\rightarrow A$ \\
    \hline
          train &   $0.15 \pm 0.07$ &    $8.47 \pm 3.33$ &   $0.27 \pm 0.08$ &   $16.82 \pm 1.86$ &   $0.14 \pm 0.16$ &   $19.28 \pm 3.97$ \\
           kite &  $-0.09 \pm 0.04$ &   $-2.67 \pm 3.35$ &  $-0.16 \pm 0.07$ &    $-7.56 \pm 3.4$ &  $-0.22 \pm 0.05$ &   $-4.89 \pm 3.78$ \\
            cow &  $-0.17 \pm 0.08$ &     $1.6 \pm 1.95$ &   $0.11 \pm 0.08$ &     $0.86 \pm 3.0$ &   $0.15 \pm 0.07$ &   $-3.58 \pm 2.55$ \\
           skis &   $0.12 \pm 0.09$ &    $0.24 \pm 2.13$ &   $0.25 \pm 0.03$ &   $-1.31 \pm 3.01$ &    $0.23 \pm 0.1$ &    $-7.5 \pm 1.76$ \\
          truck &  $-0.27 \pm 0.04$ &  $-13.33 \pm 1.68$ &  $-0.25 \pm 0.12$ &  $-10.26 \pm 1.42$ &  $-0.16 \pm 0.03$ &  $-17.95 \pm 2.46$ \\
       elephant &  $-0.58 \pm 0.15$ &    $9.09 \pm 1.25$ &  $-0.24 \pm 0.08$ &    $4.29 \pm 2.94$ &  $-0.63 \pm 0.06$ &    $1.69 \pm 3.13$ \\
           boat &   $0.03 \pm 0.05$ &     $2.8 \pm 1.79$ &  $-0.02 \pm 0.07$ &    $-0.0 \pm 2.48$ &   $0.05 \pm 0.05$ &     $3.2 \pm 3.06$ \\
        frisbee &   $0.09 \pm 0.17$ &   $-1.36 \pm 1.18$ &  $-0.14 \pm 0.22$ &     $0.7 \pm 2.45$ &  $-0.79 \pm 0.17$ &   $-6.88 \pm 2.28$ \\
       airplane &   $0.14 \pm 0.07$ &    $4.23 \pm 3.27$ &   $0.15 \pm 0.07$ &     $5.77 \pm 4.4$ &   $0.16 \pm 0.05$ &    $5.77 \pm 3.69$ \\
     motorcycle &  $-0.06 \pm 0.04$ &   $-6.35 \pm 3.94$ &   $0.01 \pm 0.06$ &   $-9.04 \pm 2.42$ &   $0.12 \pm 0.04$ &   $-1.73 \pm 6.64$ \\
      surfboard &  $-0.02 \pm 0.05$ &    $3.83 \pm 4.79$ &  $-0.06 \pm 0.07$ &    $1.74 \pm 1.36$ &  $-0.08 \pm 0.06$ &    $2.43 \pm 4.83$ \\
            tie &   $0.16 \pm 0.06$ &     $3.29 \pm 2.1$ &   $0.08 \pm 0.07$ &    $3.53 \pm 3.45$ &   $0.23 \pm 0.04$ &    $4.71 \pm 4.47$ \\
      snowboard &    $0.4 \pm 0.13$ &    $10.05 \pm 2.1$ &   $0.53 \pm 0.16$ &     $9.64 \pm 1.0$ &   $0.41 \pm 0.22$ &      $7.9 \pm 1.7$ \\
   baseball bat &   $0.46 \pm 0.04$ &   $-3.72 \pm 2.47$ &    $0.45 \pm 0.1$ &    $-3.36 \pm 1.8$ &   $0.23 \pm 0.12$ &    $1.24 \pm 2.17$ \\
 baseball glove &  $-0.03 \pm 0.05$ &   $27.06 \pm 4.12$ &  $-0.16 \pm 0.04$ &    $34.9 \pm 1.29$ &   $-0.1 \pm 0.03$ &   $36.47 \pm 0.84$ \\
     skateboard &  $-0.28 \pm 0.03$ &   $11.37 \pm 3.51$ &  $-0.34 \pm 0.08$ &   $10.98 \pm 3.86$ &  $-0.35 \pm 0.11$ &    $5.88 \pm 3.77$ \\

    \bottomrule
  \end{tabular}
  \end{table}


\end{document}